\documentclass[conference]{IEEEtran}
\pdfoutput=1
\IEEEoverridecommandlockouts
\usepackage{cite}
\usepackage{amsmath,amssymb,amsfonts}
\usepackage{algorithmic}
\usepackage{graphicx}
\usepackage{textcomp}
\usepackage{xcolor}
\usepackage[outdir=./figs/]{epstopdf}
\usepackage{graphicx}
\usepackage{subfigure}
\usepackage{cases}
\usepackage{comment}

\usepackage{color}

\usepackage{dblfloatfix}    % To enable figures at the bottom of page

\specialcomment{old}{\color{red}}{\color{black}}
\specialcomment{new}{\color{blue}}{\color{black}}

\excludecomment{old}
\specialcomment{new}{\color{black}}{\color{black}}

\usepackage{tikz}
\usepackage{fancyhdr}

\fancypagestyle{mahmood}{%
   \fancyhf{} % clear all fields
   
   \fancyhead[L]{\small \textsf{In: Proceedings of the Joint IEEE International Conference on Development and Learning and on Epigenetic Robotics (ICDL-EpiRob), Oslo, Norway, Aug. 19-22, 2019}}
}%

\makeatletter
\let\ps@IEEEtitlepagestyle\ps@mahmood
\makeatother

\newcommand\copyrighttext{%
  \footnotesize \textcopyright 2019 IEEE. Personal use of this material is permitted. Permission from IEEE must be obtained for all other uses, in any current or future media, including reprinting/republishing this material for advertising or promotional purposes, creating new collective works, for resale or redistribution to servers or lists, or reuse of any copyrighted component of this work in other works}
\newcommand\copyrightnotice{%
\begin{tikzpicture}[remember picture,overlay]
\node[anchor=south,yshift=10pt] at (current page.south) {\fbox{\parbox{\dimexpr\textwidth-\fboxsep-\fboxrule\relax}{\copyrighttext}}};
\end{tikzpicture}%
}

\newcommand\blfootnote[1]{
	\begingroup
	\renewcommand\thefootnote{}\footnote{\vspace{-0.8cm} #1}
	\addtocounter{footnote}{-1}
	\endgroup
}

\begin{document}

\title{\vspace{0.1cm} Hierarchical Control for Bipedal Locomotion using Central Pattern Generators and Neural Networks\\
\thanks{This work was supported by the German Research Foundation DFG under project CML (TRR 169).}
}

\author{\IEEEauthorblockN{Sayantan Auddy}
\IEEEauthorblockA{\textit{Intelligent and Interactive Systems} \\
\textit{University of Innsbruck}\\
Innsbruck, Austria \\
sayantan.auddy@uibk.ac.at}
\and
\IEEEauthorblockN{Sven Magg}
\IEEEauthorblockA{\textit{Knowledge Technology} \\
\textit{University of Hamburg}\\
Hamburg, Germany \\
magg@informatik.uni-hamburg.de}
\and
\IEEEauthorblockN{Stefan Wermter}
\IEEEauthorblockA{\textit{Knowledge Technology} \\
\textit{University of Hamburg}\\
Hamburg, Germany \\
wermter@informatik.uni-hamburg.de}
}

\maketitle
\blfootnote{\copyrightnotice}

\begin{abstract}
The complexity of bipedal locomotion may be attributed to the difficulty in synchronizing joint movements while at the same time achieving high-level objectives such as walking in a particular direction. Artificial central pattern generators (CPGs) can produce synchronized joint movements and have been used in the past for bipedal locomotion. However, most existing CPG-based approaches do not address the problem of high-level control explicitly.
We propose a novel hierarchical control mechanism for bipedal locomotion where an optimized CPG network is used for joint control and a neural network acts as a high-level controller for modulating the CPG network. By separating motion generation from motion modulation, the high-level controller does not need to control individual joints directly but instead can develop to achieve a higher goal using a low-dimensional control signal. The feasibility of the hierarchical controller is demonstrated through simulation experiments using the Neuro-Inspired Companion (NICO) robot. Experimental results demonstrate the controller's ability to function even without the availability of an exact robot model.
\end{abstract}


\section{INTRODUCTION}

\begin{old}
In nature, central pattern generators (CPGs) exist as neural circuits in the spinal cord of vertebrates and have been found to be responsible for the rhythmic movements of animals \cite{
Grillner1985NeuralVertebrate, 
Stein1999Neurons, 
Delcomyn1980AnimalsCPG, Arena2000CPG, Shepherd1994NeurobiologyBook}.
Artificial CPGs, constructed using coupled differential equations, mimic the behavior of natural CPGs and have a number of properties that make them a suitable choice for robot locomotion. Firstly, CPGs can produce rhythmic output without requiring rhythmic input and the output can be modulated using a few control parameters. Secondly, CPGs exhibit stable limit cycle behavior and can return to their inherent rhythmic behavior after being subject to external disturbances. Thirdly, by integrating sensory feedback into the differential equations, CPGs can achieve mutual entrainment with the natural dynamics of a robot \cite{Ijspeert2008CPGReview}.
\end{old}

\begin{new}
In nature, central pattern generators (CPGs) exist as neural circuits in the spinal cord of vertebrates and have been found to be responsible for the rhythmic movements of animals \cite{
Grillner1985NeuralVertebrate, 
Stein1999Neurons, 
Delcomyn1980AnimalsCPG, Arena2000CPG, Shepherd1994NeurobiologyBook}.
Artificial CPGs mimic the behavior of natural CPGs and have a number of properties suitable for robot locomotion. CPGs can produce rhythmic output autonomously, they exhibit stable limit cycle behavior, and they can achieve mutual entrainment with the natural dynamics of a robot \cite{Ijspeert2008CPGReview}.
\end{new}

CPGs such as the Matsuoka oscillator \cite{Matsuoka1985, Matsuoka1987} have been used in the past for bipedal locomotion \cite{Ishiguro2003Neuromodulated, Cristiano2013CPGUneven,  Cristiano2014LocomotionCPGFeedback, Cristiano2017GenerationControlCPG} by employing a network of interconnected CPGs, each of which controls an individual robot joint. Feedback pathways are incorporated to allow for some degree of adaptation and the CPG parameters are tuned by an optimization algorithm. The fitness criterion of the optimization process is designed to yield basic gait characteristics such as stability or speed. The adaptive capability of such an approach is restricted because the feedback mechanisms, which are often implemented using linear coupling terms in the CPG differential equations, have a limited capability of reacting to environmental changes. 

%\begin{figure}[!t]
%        \centerline{
%        \framebox{\parbox{0.47\textwidth}{
%        \subfigure
%        {
%            \includegraphics[width=0.45\textwidth]{figs/Architecture_overview_paper2}
%            %\label{fig:first_sub}
%        }}}}
%
%        \vspace*{1mm}        
%        
%        \centerline{
%        \framebox{\parbox{0.47\textwidth}{
%        \subfigure
%        {
%            \includegraphics[width=0.45\textwidth]{../../report/src/nico_walk_8frames_wtmpc_af1_paper.png}
%            %\label{fig:second_sub}
%        }}}}
%\caption{Top: Overview of the hierarchical bipedal controller. Bottom: Snapshots of the NICO's walking motion using the hierarchical controller.}
%\label{fig:architecture_overview}        
%\end{figure} 

\begin{figure*}[t!]
\centerline{\framebox{
\includegraphics[scale=0.3]{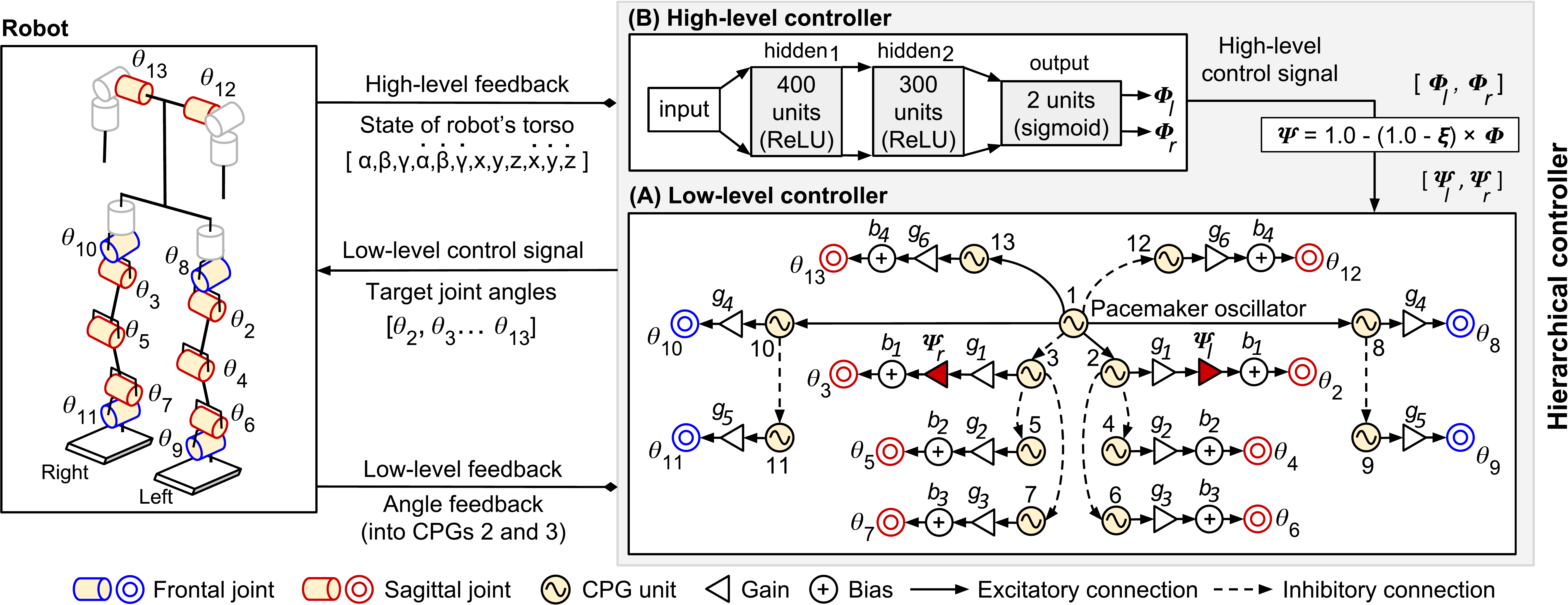}
}}
\caption{Detailed architecture. Left: Robot joints are controlled by the CPG network. Joint $\theta_i$ is controlled by CPG $i$. Right: (A) Low-level controller CPG network with 13 CPGs and their gains and biases (modified from \cite{Cristiano2014LocomotionCPGFeedback}). Apart from CPG-1 (pacemaker oscillator), each CPG controls a single joint. (B) High-level controller neural network, whose outputs $(\Phi_l, \Phi_r)$ are converted to $(\Psi_l, \Psi_r)$ for modulating the CPG network.}
\label{fig:architecture_detail}
\end{figure*}

In order to improve the adaptability and performance of a CPG-based bipedal controller, we propose a novel hierarchical architecture, in which a trainable high-level controller is added on top of a CPG network.       
The lower level of the controller, consisting of a CPG network (based on \cite{Cristiano2014LocomotionCPGFeedback}) and its feedback pathways, is used for generating target angles for the robot joints. A generalized Matsuoka oscillator \cite{Kamimura2005automaticMatsuokaModification} is used for modeling the CPGs and the CPG network parameters are found by an optimization algorithm.  The high-level controller is implemented as a feedforward neural network trained through deep reinforcement learning, whose outputs are used to regulate selected CPG network parameters. 
\color{black} %changed
This enables the high-level controller to modulate the CPG network's behavior using a low-dimensional feedback signal without needing to explicitly control every single joint.
\color{black}
%This enables the high-level controller to modulate the CPG network's behavior based on a high-level feedback signal which forms the neural network's input. 
\color{black} %changed 
On purpose, we used a robot model with minor asymmetries between the left and right sides. Together with slippage on the ground, this causes the robot to deviate laterally when only the CPG network is used for walking. 
\color{black}
As an example of simple high-level control, we demonstrate how the neural network can learn to adjust to the robot model's inaccuracies and make it walk straight.
%The presence of minor discrepancies in the robot model and slippage between the feet and the ground cause the robot to deviate laterally when only the CPG network is used for walking. As an example of high-level control, we demonstrate through experimental results how the high-level controller can learn to rectify the undesired turning behavior of the robot and make it walk straight.

%is able to make the robot walk straight even in the presence of imperfections in its structure.
 
%The remainder of this paper is organized as follows. In Section \ref{sec:related_work}, related work is presented followed by a detailed description of the controller in Section \ref{sec:architecture}. Experimental results are presented in Section \ref{sec:experiments} and finally, concluding remarks and future work are presented in Section \ref{sec:conclusions}.

\section{RELATED WORK}
\label{sec:related_work}

Methods based on zero moment point (ZMP) \cite{Kajita2003PreviewControlZMP, Kajita20013DLIPM, Kamogawa2013ZMP, Yu2016disturbanceZMP} have been a common choice for bipedal locomotion in robots. A criticism of such methods is that the reference trajectories generated using simplified models constrain the robot's movements, resulting in an unnatural gait with bent knees \cite{Missura2015PushRecovery} and high energy consumption \cite{Collins2005PassiveDynamics}. McGeer's \cite{Mcgeer1990PassiveWalking} passive walking machines and Collins et al.'s \cite{Collins2005PassiveDynamics} minimally actuated robots showed that if the body's natural dynamics is utilized, bipedal locomotion is possible even without complicated approaches.

CPG-based bipedal methods seek to utilize the ability of CPGs to entrain with the body dynamics of the robot. Taga et al. \cite{Taga1991SelfOrganizedBiped} used a network of coupled Matsuoka oscillators to control a 5-link planar biped in simulation. They showed that stable and flexible locomotion could be generated by entrainment between the rhythmic activities of the coupled CPGs and the movements of the mechanical structure. Ishiguro et al. \cite{Ishiguro2003Neuromodulated} used a network of Matsuoka oscillators to control a 3D bipedal robot in simulation. An artificial neuromodulation mechanism was used for modulating the CPGs and the parameters were set by a genetic algorithm. Endo et al. \cite{Endo2008HumanoidPoilicyGradientRL} used a CPG-based controller to achieve bipedal locomotion in a physical humanoid robot. The walking motion was broken up into a stepping-in-place motion and a propulsive motion. The feedback pathways for the propulsive motion were learned using a policy-gradient based method. The developed feedback controller showed stable walking behavior in a physical environment. Cristiano et al. \cite{Cristiano2014LocomotionCPGFeedback} implemented a CPG network, composed of Matsuoka oscillators, for controlling the Nao robot's walking behavior in simulation as well as in the real world. One of the CPGs, designated as the pacemaker oscillator was used for generating the master signal for driving the CPG network. CPG network parameters were optimized by a genetic algorithm.

Most of these CPG-based methods use some form of feedback control, but the feedback parameters are fixed after optimization. Introducing a trainable high-level controller, which can modulate the CPG network, will lead to greater generalization abilities and a more effective form of control, 
which motivates the approach followed in this paper.
%Hence, in this paper a neural network is trained using reinforcement learning and used to modulate selected CPG network parameters.

\section{ARCHITECTURE}
\label{sec:architecture}

The architecture of the hierarchical controller (Fig. \ref{fig:architecture_detail}) depicts three distinct components: the robot, the low-level controller (CPG network) and the high-level controller (neural network).
A simulated NICO humanoid robot \cite{Nicopaper} has been used to develop our bipedal locomotion approach. The NICO is 101cm tall, weighs 7kg and has 30 DoF (degrees-of-freedom) in total. Each leg has 3 DoF in the hip, 1 DoF in the knee and 2 DoF in the ankle. The proposed controller controls 10 sagittal and frontal joints in the legs and 2 sagittal joints in the shoulders (since arm swing increases walking stability).

\subsection{Low-level Controller} \label{subsec:lowlevel}

For the low-level controller, we adapted the CPG network from \cite{Cristiano2014LocomotionCPGFeedback}. 
%First the basic CPG network is briefly described and then the modifications made by us are detailed. 
The CPG network consists of 13 generalized Matsuoka oscillators \cite{Kamimura2005automaticMatsuokaModification}, interconnected using either an excitatory connection (weight = $+1$) causing the connected CPGs to oscillate in-phase, or an inhibitory connection (weight = $-1$) causing anti-phase oscillations. One pacemaker oscillator is responsible for driving the other CPGs and maintaining the phase relationships between them. The other 12 CPGs control separate robot joints. Empirically chosen, identical parameters are used for all the CPGs. Each CPG output is varied by multiplication with a gain term (for changing the amplitude) and addition with a bias term (for changing the mean position). Gains and biases are used symmetrically across the left and right sides of the CPG network, to produce symmetrical movement. Biases are omitted for frontal joints since their mean position should be zero. The configuration of CPG connections (shown in Fig. \ref{fig:architecture_detail}) results in limb movements similar to that in humans, e.g. contralateral sagittal hip joints move in anti-phase but each sagittal hip joint moves in-phase with the contralateral sagittal shoulder joint.

We modified the CPG network of \cite{Cristiano2014LocomotionCPGFeedback} by using a different feedback mechanism. Instead of the phase reset feedback, wherein the pacemaker oscillator's phase is reset based on the pattern of foot-ground contact, we integrated the feedback mechanism from \cite{Ishiguro2003Neuromodulated} where the actual angular position of each sagittal hip joint is multiplied by a weight $k$ and then fed back into the corresponding CPG. 
%Based on detailed experiments \cite{AuddyMscThesis}, we found that this form of feedback performs better than phase reset, in terms of stability and distance covered.
In experiments in which we tested each feedback mechanism for 300 trials each (Sec. 5.2 in \cite{AuddyMscThesis}), we found that using the angular feedback mechanism resulted in a better gait in terms of distance and stability.

%The CPG network is designed to produce symmetrical movement. 
%If the robot deviates from a straight line, it would need the ability to turn while walking, to correct its course. 
To enable the robot to turn, we introduced two additional gains $\Psi_l$ and $\Psi_r$ (red triangles in Fig. \ref{fig:architecture_detail}), for the left and right sagittal hip CPGs respectively. By using unequal values for $\Psi_l$ and $\Psi_r$, the amplitude of the two sagittal hip joints can differ, causing unequal stride lengths resulting in a small turn. The values of $\Psi_l$ and $\Psi_r$ are set by the high-level controller. 

Each CPG unit is a Matsuoka oscillator consisting of an extensor and a flexor neuron whose behaviors are governed by (\ref{eq:mat1}) and (\ref{eq:mat2}), respectively (generalized Matsuoka equations \cite{Kamimura2005automaticMatsuokaModification, Cristiano2014LocomotionCPGFeedback}). Subscript $i$ denotes a CPG unit, and superscripts $e$ and $f$ denote the extensor and flexor neurons respectively.
\begin{equation}
\label{eq:mat1}
\begin{aligned}
\tau_0 \kappa \dot{u}_{i}^e &= -u_{i}^e -  w_0 y_{i}^f - \beta v_{i}^e + u_t + f_{i}^e + s_{i}^e  \\
\tau'_0 \kappa \dot{v}_{i}^e &= -v_{i}^e + y_{i}^e \\
&\text{where }y_{i}^e = max(0,u_{i}^e)\text{ and }i=1,...,num
\end{aligned} \\
\end{equation}
\begin{equation}
\label{eq:mat2}
\begin{aligned}
\tau_0 \kappa \dot{u}_{i}^f &= -u_{i}^f -  w_0 y_{i}^e - \beta v_{i}^f + u_t + f_{i}^f + s_{i}^f  \\
\tau'_0 \kappa \dot{v}_{i}^f &= -v_{i}^f + y_{i}^f  \\
&\text{where }y_{i}^f = max(0,u_{i}^f)\text{ and  }i=1,...,num
\end{aligned}\\
\end{equation} 
State variables $u_{i}^e$, $u_{i}^f$ control the discharge rate and $v_{i}^e$, $v_{i}^f$ control the self-inhibition of the extensor and flexor neurons respectively. $\tau_0$ and $\tau'_0$ are the time constants for the rate of discharge and adaptation, respectively. $\kappa$ is a parameter which modulates the frequency of the CPG. $y_{i}^e$ and $y_{i}^f$ are the activations of the extensor and flexor neurons. $\beta$ and $w_0$ are the constants of self- and mutual-inhibition respectively. $u_t$ is the tonic input and $num$ is the number of CPG units in the network. $f_{i}^e$ and $f_{i}^f$ are feedback terms, which are non-zero only for the sagittal hip oscillators, for which $f_{i}^e=k\theta_i'$ and $f_{i}^f=-k\theta_i'$, where $\theta_i'$ is the actual angle of the joint controlled by CPG $i$ and $k$ is the feedback weight.  $s_{i}^e=w_{ij}u_{j}^e$, $s_{i}^f=w_{ij}u_{j}^f$ represent the interaction between connected CPGs $i$ and $j$ ($w_{ij}$ is the connection weight). Output $o_i$ of CPG $i$ is obtained by $o_i=-y_{i}^e+y_{i}^f$, which is multiplied by a gain and added with a bias (for sagittal CPGs), as shown in Fig. \ref{fig:architecture_detail}.

\subsection{High-level Controller} \label{subsec:highlevel}

The CPG network is designed to produce symmetrical joint movement across the left and right sides of the robot. However, the presence of structural inconsistencies in the legs, and slippage with the ground cause the robot to deviate from a straight trajectory when the low-level controller is used in isolation. 
%With a perfect model this deviation would be eliminated in simulation, but in the real world, the robot would still be subject to errors. 
In this paper, we use the high-level controller to rectify the turning behavior, but its function can be easily extended to other high-level objectives as well.
%The high-level controller monitors the movement of the robot's torso and modulates the CPG network in order to achieve an overall goal. It does not control or monitor individual joints directly. To demonstrate the concept, a simple form of high-level control is used in this paper. Due to minor inconsistencies in the legs and slippage during walking, the robot deviates from a straight line trajectory when it walks for an extended duration of time. With a perfect model this deviation would be eliminated in simulation, but in the real world, the robot would still be subject to errors. It is crucial for the controller to function even in the presence of systematic and non-systematic errors. 
Any lateral deviation is minimized by adjusting the stride lengths of the left and right feet (by varying the gain parameters $\Psi_l$, $\Psi_r$). The target angles $\theta_2$ and $\theta_3$ for the left and right sagittal hip joints are given by (\ref{eq:high_level_turn}), where $o_2$ and $o_3$ are the outputs of the corresponding CPGs, and $g_1$ and $b_1$ are the gain and bias applied to the CPG outputs.
\begin{equation}
\label{eq:high_level_turn}
\begin{aligned}
\theta_2&=o_2\Psi_lg_1+b_1 \\
\theta_3&=o_3\Psi_rg_1+b_1
%\theta_2=o_2\Psi_lg_1+b_1 \mathrm{\text{\ \ \ and\ \ \ }} \theta_3=o_3\Psi_rg_1+b_1
\end{aligned}
\end{equation}

The high-level controller is implemented as a fully connected, feedforward neural network with two hidden layers containing 400 and 300 ReLU units respectively (the structure of the hidden layers is based on the actor network used in \cite{Lillicrap2015DDPG}). The output layer consists of 2 sigmoid units. The input to the network is the vector $[\alpha, \beta, \gamma, \dot{\alpha}, \dot{\beta}, \dot{\gamma}, x, y, z, \dot{x}, \dot{y}, \dot{z}]$, consisting of the angular position $(\alpha, \beta, \gamma)$ and velocity $(\dot{\alpha}, \dot{\beta}, \dot{\gamma})$, and the Cartesian position $(x, y, z)$ and velocity $(\dot{x}, \dot{y}, \dot{z})$ of the robot's torso in the three dimensions. The outputs of the network ($\Phi_l$, $\Phi_r$) are used to derive the gains ($\Psi_l$, $\Psi_r$), according to (\ref{eq:phi_psi_conv}), where $\xi \in [0.0, 1.0]$ is a parameter which controls how much influence the high-level controller can exert over the low-level controller.
\begin{equation}
\label{eq:phi_psi_conv}
\begin{aligned}
\Psi_l&=1.0-(1.0-\xi)\Phi_l \\
\Psi_r&=1.0-(1.0-\xi)\Phi_r
%\Psi_l=1.0-(1.0-\xi)\Phi_l\mathrm{\text{ and }}\Psi_r=1.0-(1.0-\xi)\Phi_r
\end{aligned}
\end{equation}
When $\xi=1.0$, $\Psi_l$ and $\Psi_r$ evaluate to 1.0, irrespective of the neural network's output ($\Phi_l$ and $\Phi_r$). In this case, the high-level controller has no influence over the CPG network's behavior because the angles for the sagittal hip joints (\ref{eq:high_level_turn}), computed by the CPG network, remain unaffected. When $\xi=0.0$, $\Psi_l$ and $\Psi_r$ will be fully dependent on $\Phi_l$ and $\Phi_r$ respectively, and hence, the high-level controller can influence the CPG network to a great extent.

The inverse relationship between $(\Psi_l,\Psi_r)$ and $(\Phi_l,\Phi_r)$ in (\ref{eq:phi_psi_conv}) is necessary because the initialization of the final layer weights and biases of the neural network (described in Section \ref{subsec:train}) is such that the network outputs are near zero in the initial stages of training. If $\Phi_l$ and $\Phi_r$ (both having near-zero values) are used in place of $\Psi_l$ and $\Psi_r$ in (\ref{eq:high_level_turn}), the angles $\theta_2$ and $\theta_3$ would not show any oscillatory behavior about the bias position. Hence the robot would not be exhibiting any forward motion and training the neural network would not be possible.
%Thus, $\Phi_l$ and $\Phi_r$ are interpreted as parameters which suppress the amplitude of the sagittal hip joints. Since $\Psi_l$ and $\Psi_r \in [0.0, 1.0]$, a value of 0.0 will denote no reduction in the amplitude and 1.0 will indicate that the amplitude of the respective joint is to be maximally reduced. 

Using this setup, the low-level controller can be optimized to produce a basic stable gait without bothering about the lateral deviation. The high-level controller can then be trained to minimize the lateral deviation using the parameters $\Psi_l$ and $\Psi_r$, without having to deal with the 12 joints individually.
\color{black} %changed
Controlling 12 joints would have required 12 outputs from the high-level neural network controller which would have significantly increased the complexity of the training process.
\color{black}

\section{EXPERIMENTS AND RESULTS}
\label{sec:experiments}

The hierarchical controller was constructed in two phases. In the first phase, the neural network was omitted, and the CPG network was optimized by a genetic algorithm (GA). In the second phase, the neural network was added on top of the optimized CPG network and trained using the deep deterministic policy gradient (DDPG) algorithm \cite{Lillicrap2015DDPG}. After training, the entire controller was tested. All experiments were conducted using  an Intel Core i5-6500 CPU and 16GB RAM. Simulations were carried out using the V-REP simulator. %\cite{vrep2013}.

\subsection{Low-level Controller Optimization}
\label{subsec:opti}

Instead of optimizing the internal parameters (described in section \ref{subsec:lowlevel}) of all the CPGs, first, the pacemaker oscillator's parameters were set empirically and replicated in the other CPGs. Then the gains, biases, the frequency controlling parameter $\kappa$ and the low-level feedback weight $k$ were optimized. A real-valued chromosome with 12 elements was used: $[\kappa, g_1, g_2, g_3, g_4, g_5, g_6, b_1, b_2, b_3, b_4, k]$. The gains $g_1,..,g_6$ and biases $b_1,..,b_4$ were used symmetrically across the left and right sides of the CPG network as shown in Fig. \ref{fig:architecture_detail}. 
%This reduces the search space for the optimization problem, but also does not enable the CPG network to account for model inconsistencies or errors between the left and right sides. 
This significantly reduces the optimization search space but does not allow the CPG network to address asymmetries through errors in the robot model, which are instead handled by the high-level controller.
To bound the search space, the limits for the chromosome values were set as: $\kappa\in[0.2,1.0]$, $g_1,..,g_6 \in [0.01,1.0]$, $b_1 \in [-0.06,0.0]$, $b_2 \in [0.0,0.5]$, $b_3 \in [-0.5,0.0]$, $b_4 \in [0.0,1.0]$, $k \in [-2.5,2.5]$. The bias limits were set so that the mean position of the sagittal joints resulted in a stable pose. The other limits were set empirically. 

The internal CPG parameters were set according to \cite{Cristiano2014LocomotionCPGFeedback} ($\tau_0=0.28$, $\tau_0'=0.4977$, $\beta=2.5$, $w_0=2.2829$, $u_t=0.4111$). A GA with a population size of 200 was run for 30 generations. We used tournament selection with a tournament size of 3, and a 2-point crossover with a probability of 80\% with random crossing points. Mutation to a gene value was performed by adding a small number, drawn randomly from a Gaussian distribution (mean=0.0, variance=$10^{-4}$) with probability 10\% for a chromosome and 5\% for a gene. For evaluating fitness, the CPG network was initialized with the values in the chromosome and the robot was allowed to walk for a maximum of 20 seconds or until it fell. Afterwards, the forward distance ($d_x$) and the time for which the robot was upright ($t_{up}$) were used to calculate the fitness using (\ref{eq:ga_fitness}).
\begin{equation}
\label{eq:ga_fitness}
fitness = d_x + 0.5 \times t_{up}
\end{equation}
We used a simple fitness function with no penalty for lateral deviation
to simplify the optimization process.
%and because the focus during optimization was only on finding a stable walk.
%since the focus of the optimization process was only on obtaining CPG network parameters which resulted in a stable walk. 
The genetic algorithm was executed multiple times, with similar results. 
%The run with the best results is shown in Fig. \ref{fig:ga_plot}. 
The best solution obtained for a stable walk was [$\kappa$ = 0.3178, $g_1$ = 0.3777, $g_2$ = 0.0234, $g_3$ = 0.0132, $g_4$ = 0.4567, $g_5$ = 0.2019, $g_6$ = 0.3309, $b_1$ = -0.0519, $b_2$ = 0.0963, $b_3$ = -0.1156, $b_4$ = 0.4814, $k$ = 1.5364].  
%\color{black} %changed
%The total time taken by the optimization process for 30 generations was approximately 23.5 hours.
%\color{black}

%\begin{figure}[htbp]
%\centerline{
%\includegraphics[scale=0.19]{figs/ga_plot_paper-eps-converted-to}}
%\caption{Genetic algorithm results. Left: fitness, right: maximum distance.}
%\label{fig:ga_plot}
%\end{figure}

% Need to place this one page before where the figure should appear
\setcounter{figure}{2} % Due to problem in numbering, arising from the placement of a wide image at the bottom of a page
\begin{figure*}[!b]
\centerline{
\includegraphics[scale=0.2]{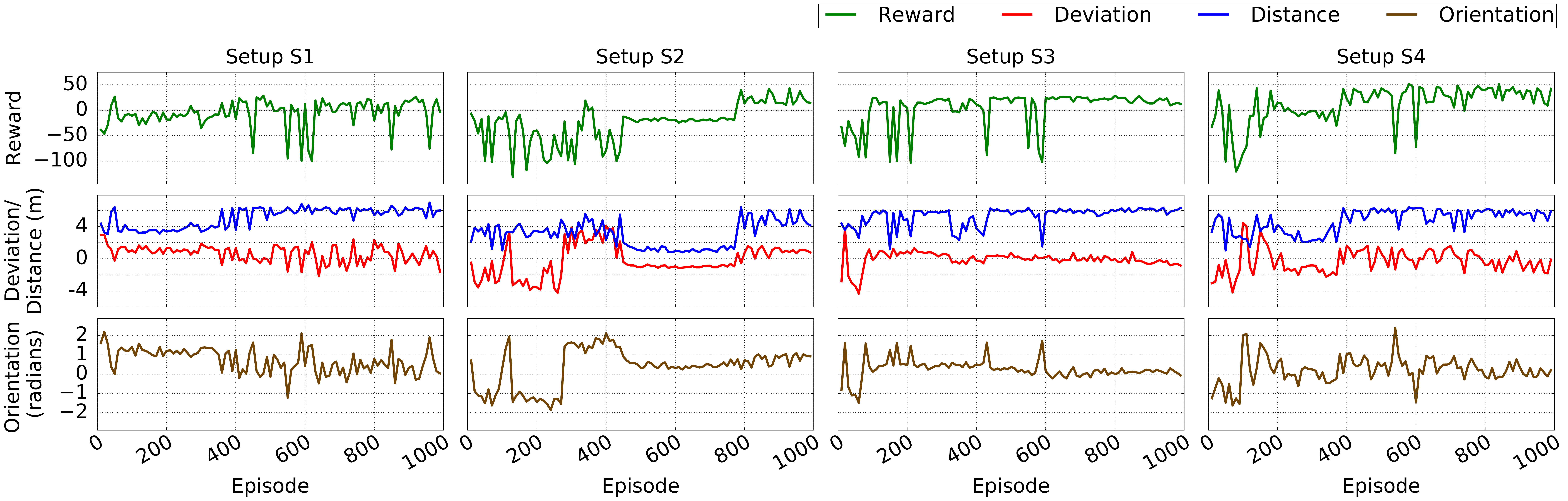}
}
\caption{Results of training the high-level controller using the setups S1-S4.}
\label{fig:rl_train}
\end{figure*}

\subsection{High-level Controller Training and Testing}
\label{subsec:train}

The neural network (high-level controller) was trained based on the DDPG algorithm \cite{Lillicrap2015DDPG} which is an off-policy, model-free, actor-critic reinforcement learning algorithm, capable of handling continuous state and action spaces. In the reinforcement learning setup, the high-level controller was the actor, responsible for implementing the policy function. At each timestep $t$, the actor network took as input a state $s_t \in \mathbb{R}^{12}$ (high-level feedback $[\alpha, \beta, \gamma, \dot{\alpha}, \dot{\beta}, \dot{\gamma}, x, y, z, \dot{x}, \dot{y}, \dot{z}]$ in Fig. \ref{fig:architecture_detail}) and produced an action $a_t \in \mathbb{R}^2$ (high-level control signal $[\Phi_l, \Phi_r]$). As a consequence, a reward $r_t \in \mathbb{R}$ and a new state $s_{t+1}$ were generated. The transition $(s_t, a_t, r_t, s_{t+1})$ was stored in a first-in-first-out cache called the \textit{replay buffer}. A separate critic network, shown in Fig. \ref{fig:critic}, estimated the action-values (Q-values). DDPG makes use of \textit{target actor} and \textit{critic networks} for improving the stability of the learning process. Minibatches from the replay buffer and the outputs of the target networks were used for updating the critic network. The outputs of the critic network were then used for sampling the \textit{policy gradient}, which was used for updating the weights and biases of the actor network.
\setcounter{figure}{1} % Due to problem in numbering, arising from the placement of a wide image at the bottom of a page

\begin{figure}[htbp]
\centerline{
\framebox{
\includegraphics[scale=0.24]{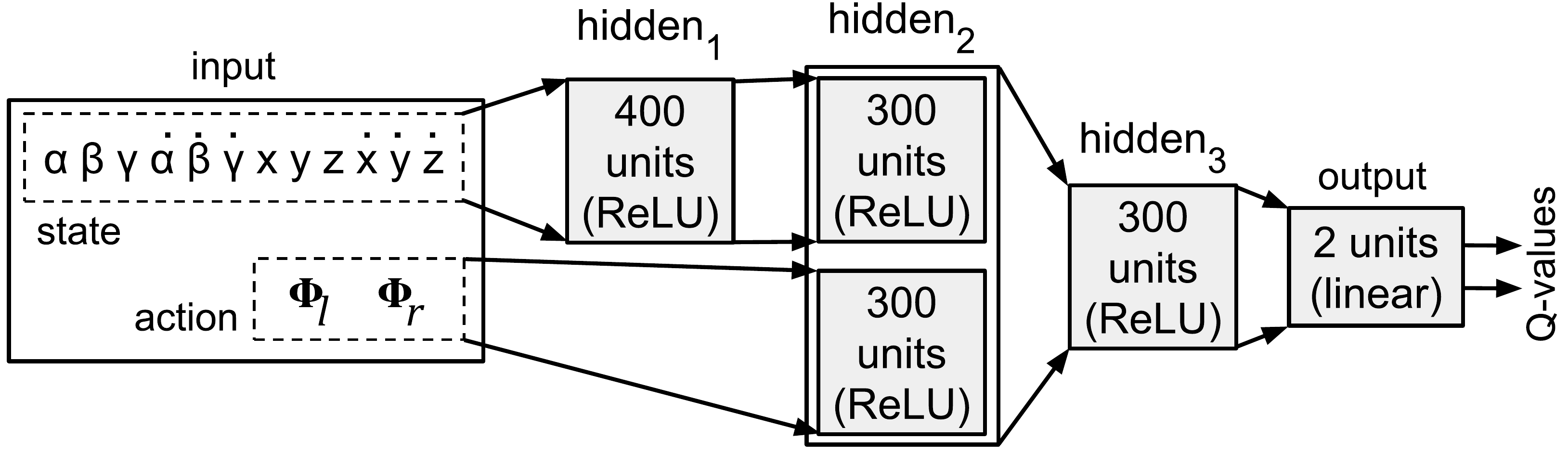}
}}
\caption{Critic network. The 2 blocks in hidden$_2$ are merged by summation.}
\label{fig:critic}
\end{figure}

During training, the high-level neural network modulated the optimized CPG network to make the robot walk. 
%The high-level controller was trained for 1000 episodes. 
In each training episode, the robot started at the origin of the world frame, facing in the $x$-direction. Each episode lasted for 40 seconds or until the robot fell down. After every second during an episode, the reward $r_t$ was calculated using (\ref{eq:rl_reward_function}).
\begin{equation}
\label{eq:rl_reward_function}
\begin{aligned}
r_t=
\begin{cases} 
-100  &\text{if robot fell}\\
-\zeta_{dev} \cdot |d_y| + \zeta_{dist} \cdot d_x - \zeta_\gamma \cdot |\gamma| &\text{otherwise}\\
\end{cases}\\
\end{aligned}
\end{equation}

The reward function was designed to favor a long forward distance ($d_x$) and to penalize any lateral deviation ($d_y$) or change in torso orientation (for the ideal straight walk, the torso orientation $\gamma$ about the world $z$-axis would always be 0 radians). The terms $\zeta_{dev},\zeta_{dist}$ and $\zeta_\gamma$ were weights assigned to the deviation, distance and orientation components, respectively. Together with $\xi$, the factor controlling the influence of the high-level controller (used in (\ref{eq:phi_psi_conv})), $\zeta_{dev},\zeta_{dist}$ and $\zeta_\gamma$ formed the 4 hyperparameters specific to the high-level controller. Different values for these hyperparameters were set to create the setups shown in Table \ref{tab:rl_ddpg_setups}.
\begin{table}[h]
\caption{Hyperparameters for training the High-Level controller}
\label{tab:rl_ddpg_setups}
\begin{center}
\begin{tabular}{c|c|c|c|c}
\hline
\multicolumn{1}{c|}{Setup} & $\zeta_{dev}$ & $\zeta_{dist}$ & $\zeta_{\gamma}$ & $\xi$ \\ \hline \hline
$S1$                   & 1.0           & 0.5            & 1.0              & 0.1   \\ \hline
$S2$                   & 1.0           & 0.5            & 1.0              & 0.4   \\ \hline
$S3$                   & 1.0           & 0.3            & 1.0              & 0.1   \\ \hline
$S4$                   & 1.0           & 0.3            & 1.0              & 0.4   \\ \hline
\end{tabular}
\end{center}
\end{table}

We used sigmoid and linear activations in the output layer of the actor and critic networks, respectively, and used a replay buffer size of $10^5$. All other DDPG-specific hyperparameters were set as per \cite{Lillicrap2015DDPG}. The final layer weights and biases of the actor and critic networks were initialized from a uniform distribution $[-3\times 10^{-3}, 3\times 10^{-3}]$, so that the initial outputs of both networks would be close to zero. The weights and biases of all the other layers were initialized from the uniform distribution $[-\frac{1}{\sqrt{f}},\frac{1}{\sqrt{f}}]$, where $f$ was the fan-in of the layer.

For each setup in Table \ref{tab:rl_ddpg_setups}, the high-level controller was trained for 1000 episodes. After every 10 episodes, the performance of the networks was evaluated. During these test episodes, the exploratory noise used in DDPG was switched off and network updates were not made. The reward, distance, deviation, and torso orientation were recorded. These results are shown in Fig. \ref{fig:rl_train}. It can be seen that by the 1000$^{th}$ episode the high-level controller for all the setups learned to maximize the distance and minimize the deviation and change in orientation. Towards the end, the robot was also more stable, since large negative rewards rarely occurred. Overall, $S3$'s results were the best since the distance stabilized around the 6m mark and the deviation and orientation stayed very close to 0 from episode 600 onwards.
%\color{black} %changed
%The duration of the training was around 17.5 hours for each setup.
%\color{black}

Once the training was complete, the trained high-level neural network for each setup was tested for 100 episodes, each of 40s duration. The distance, deviation and orientation were measured at the end of each episode. A control setup $S0$ was created by using only the optimized CPG network without any high-level controller. 
\color{black} %changed
To perform a comprehensive comparison, we also created 2 additional setups ($L1$, and $L2$) in which a simple linear feedback controller was used to modulate the CPG network by tracking the deviation of the robot from a straight trajectory. For modulating the CPG network, the linear controller used (\ref{eq:linear_control}) for setting the values of $\Phi_l$ and $\Phi_r$.

\begin{equation}
\label{eq:linear_control}
\begin{aligned}
\Big( \Phi_l,\Phi_r \Big)=
\begin{cases} 
\big( bound(\mathcal{G} \times |d_y|), 0 \big)  & d_y > 0\\
\big( 0, bound(\mathcal{G} \times |d_y|) \big)  & d_y < 0\\
\end{cases}\\
\end{aligned}
\end{equation}

In (\ref{eq:linear_control}), $\mathcal{G}$ is the gain of the linear controller and $bound(x)$ restricts $x$ within the range $[0,1]$. Since the linear controller acts as the high-level controller, (\ref{eq:phi_psi_conv}) is used to convert $\Phi_{l\vert r}$ to $\Psi_{l\vert r}$. The hyperparameters for $L1$ were $\xi=0.1, \mathcal{G}=0.2$  and those for $L2$ were     $\xi=0.1, \mathcal{G}=0.4$. These values were empirically determined.
The control setup $S0$ and the linear controller setups $L1$ and $L2$ were also tested for 100 episodes each. The performance of all the setups is shown in Fig. \ref{fig:rl_test}.
 
\setcounter{figure}{3} % Due to problem in numbering, arising from the placement of a wide image at the bottom of a page

\color{black}
\begin{figure}[htbp]
\centering{
\includegraphics[width=0.47\textwidth]{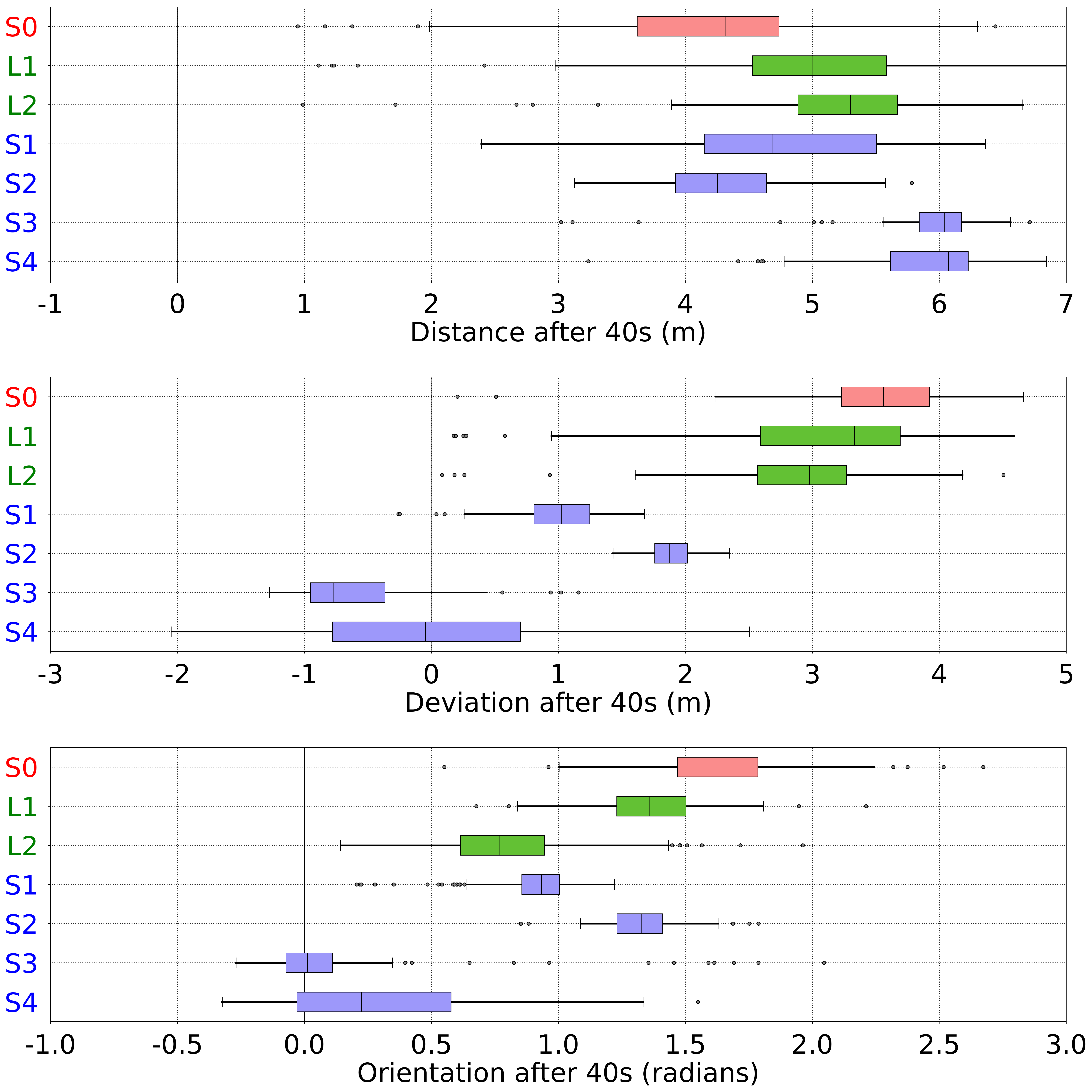}
}
\caption{Results for 100 test episodes, each with a maximum duration of 40s. Top: distance, middle: deviation, bottom: orientation. 
}
\label{fig:rl_test}
\end{figure}

\begin{figure}[!h]
\centering{
\includegraphics[width=0.5\textwidth]{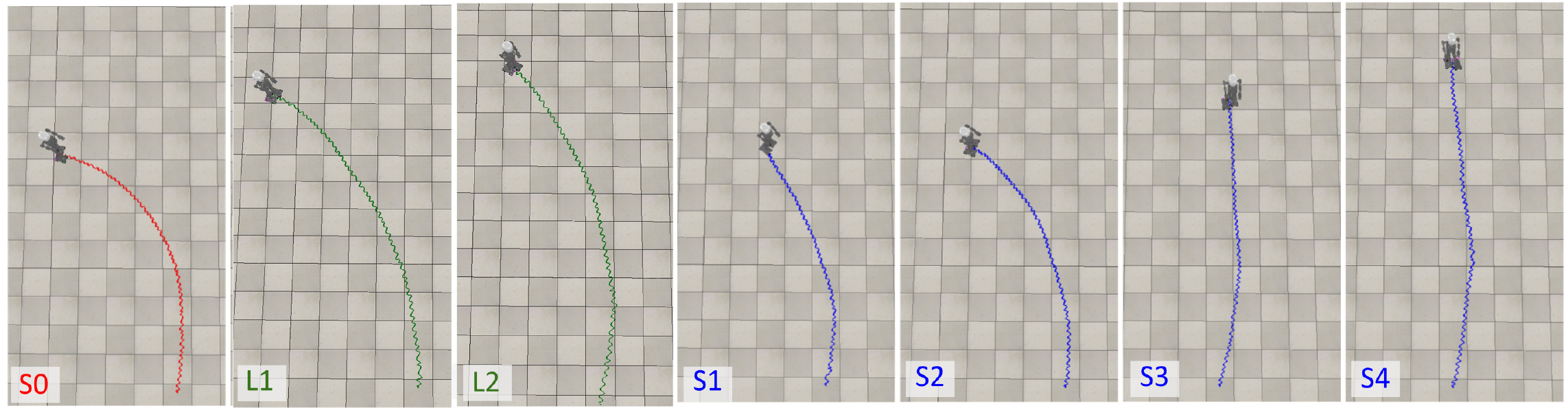}
\label{fig_first_case}}
\caption{Example trajectories for the different setups: S0, L1, L2, S1, S2, S3, S4 (left to right).}
\label{fig:paths}
\end{figure}

Fig. \ref{fig:rl_test} (top) shows that the trained high-level controller was able to increase the forward distance walked by the robot, especially for the setups $S3$ and $S4$, for which the median distances are much higher than for the control setup $S0$.
\color{black} %changed
For $L1$ and $L2$ the median distance was better than $S0$ but considerably less than $S3$ and $S4$.
\color{black}
The effect of the 
\color{black} %changed 
neural network-based 
\color{black} 
high-level controller is also evident in Fig. \ref{fig:rl_test} (middle) and (bottom), where the median deviation and orientation for $S3$ and $S4$ are very close to the ideal position of zero. The results for $S1$ and $S2$ are also better than $S0$ but not as good as $S3$ and $S4$.
\color{black} %changed
The linear feedback controller setups $L1$ and $L2$ also performed better than the standalone CPG ($S0$) but were not as effective in reducing the deviation and torso rotation as the neural network-based controller.
\color{black}
It can be seen from Fig. \ref{fig:paths}, that for $S3$ and $S4$, the robot's trajectory was straighter and hence also longer than the other setups.
The superior performance of $S3$ and $S4$ may be attributed to the fact that compared to the other 
\color{black} %changed 
neural network-based 
\color{black} 
setups, $S3$ and $S4$ assigned a lower weight to the forward distance in the reward calculation, and thereby learned to pay more attention to the undesirable behaviors of lateral deviation and change in orientation. Additionally, for $S3$, the influence of the high-level controller was more (due to a lower value of $\xi$ than $S4$), which may have contributed to the low variability in its performance compared to $S4$. 
\color{black} %changed
From Fig. \ref{fig:rl_test}, it is also evident that the simple linear equation used in (\ref{eq:linear_control}) was inadequate in dealing with the asymmetry of the robot and the ground slippage, and was not as capable as the neural network in achieving the high-level objective. The robot's walk for the different setups can be viewed at \texttt{https://youtu.be/4c64rKhj72E}.
\color{black}

\section{CONCLUSIONS}
\label{sec:conclusions}

Although the optimized CPG network can produce a stable walk, errors in the robot model or noise in the actuation or the environment can always lead to deviations.
The 
\color{black} %changed
neural network-based 
\color{black}
 high-level controller is able to deal with this problem effectively by using a simple 2-dimensional control signal 
\color{black} %changed
and without having to explicitly control every individual joint. 
\color{black}
By using a relatively simple high-level objective of walking straight, we showed that a high-level 
\color{black} %changed
neural network 
\color{black}
controller can be used to improve the performance of a CPG network for bipedal locomotion.
The same approach can be extended in the future for more complex high-level goals, such as maintaining balance while walking on uneven or sloped surfaces. This can be achieved by letting the high-level controller modulate the bias position of joints, which would affect the tilt of the robot's body. The general approach can be implemented using a different robot, or by using different CPG network configurations or neural network architectures for the high-level controller. Also it is possible to train the high-level controller for multiple high-level objectives together by designing the reward function accordingly. 
\color{black} %changed
We will also investigate the effects of using a simpler state representation and the effects of sensory noise and feedback delays.
\color{black}
%Confirming the advantages of our approach on a real NICO robot is our next step, but the presented results have already shown the positive impact of this approach on bipedal locomotion in an uncertain environment.
Confirming the advantages of our approach on the real NICO robot is our next step, but the presented results already show the promise of this biologically-inspired, developmental approach for achieving bipedal walking in uncertain environments.

\addtolength{\textheight}{-11.0cm} %-12cm   % This command serves to balance the column lengths

%%%%%%%%%%%%%%%%%%%%%%%%%%%%%%%%%%%%%%%%%%%%%%%%%%%%%%%%%%%%%%%%%%%%%%%%%%%%%%%%

%%%%%%%%%%%%%%%%%%%%%%%%%%%%%%%%%%%%%%%%%%%%%%%%%%%%%%%%%%%%%%%%%%%%%%%%%%%%%%%%

%%%%%%%%%%%%%%%%%%%%%%%%%%%%%%%%%%%%%%%%%%%%%%%%%%%%%%%%%%%%%%%%%%%%%%%%%%%%%%%%
%\section*{APPENDIX}

%\section*{ACKNOWLEDGMENT}

%%%%%%%%%%%%%%%%%%%%%%%%%%%%%%%%%%%%%%%%%%%%%%%%%%%%%%%%%%%%%%%%%%%%%%%%%%%%%%%%
\bibliographystyle{IEEEtran}
\bibliography{auddy_icdl_2019}

\end{document}